\documentclass[10pt,twocolumn,letterpaper]{article}

\usepackage[accsupp]{axessibility}
\usepackage{iccv}
\usepackage{times}
\usepackage{epsfig}
\usepackage{graphicx}
\usepackage{amsmath}
\usepackage{amssymb}

\usepackage{multirow}
\usepackage{booktabs}
\usepackage{xcolor}  
\usepackage{colortbl}
\definecolor{mygray}{gray}{.9}

\usepackage{arydshln}
\usepackage{makecell}

\usepackage[pagebackref=true,breaklinks=true,letterpaper=true,colorlinks,bookmarks=false]{hyperref}

\iccvfinalcopy 

\def\iccvPaperID{1568} 

\ificcvfinal\fi

\begin{document}

\title{UATVR: Uncertainty-Adaptive Text-Video Retrieval}

\author{
Bo Fang$^{1*}$ \qquad Wenhao Wu$^{2,3*}$\qquad Chang Liu$^{4*}$\qquad Yu Zhou$^{1\dag}$ \qquad Yuxin Song$^{3}$\\  
Weiping Wang$^{1}$\qquad Xiangbo Shu$^{5}$\qquad Xiangyang Ji$^{4}$\qquad Jingdong Wang$^{3}$\\
$^{1}$Institute of Information Engineering, Chinese Academy of Sciences \quad  $^{2}$The University of Sydney \\
$^{3}$Baidu Inc. \quad $^4$Tsinghua University 
\quad $^5$Nanjing University of Science and Technology\\
{\tt\small \{fangbo,zhouyu,wangweiping\}@iie.ac.cn,
\{songyuxin02,wangjingdong\}@baidu.com,
}
\\
{\tt\small \{liuchang2022,xyji\}@tsinghua.edu.cn,
wenhao.wu@sydney.edu.au,
shuxb@njust.edu.cn
}
}

\maketitle
\ificcvfinal\thispagestyle{empty}\fi

\begin{abstract}
With the explosive growth of web videos and emerging large-scale vision-language pre-training models, e.g., CLIP, retrieving videos of interest with text instructions has attracted increasing attention.
A common practice is to transfer text-video pairs to the same embedding space and craft cross-modal interactions with certain entities in specific granularities for semantic correspondence.
Unfortunately, the intrinsic uncertainties of optimal entity combinations in appropriate granularities for cross-modal queries are understudied, which is especially critical for modalities with hierarchical semantics, e.g., video, text, etc. 
In this paper, we propose an Uncertainty-Adaptive Text-Video Retrieval approach, termed UATVR, which models each look-up as a distribution matching procedure.
Concretely, we add additional learnable tokens in the encoders to adaptively aggregate multi-grained semantics for flexible high-level reasoning.
In the refined embedding space, we represent text-video pairs as probabilistic distributions where prototypes are sampled for matching evaluation.
Comprehensive experiments on four benchmarks justify the superiority of our UATVR, which achieves new state-of-the-art results on MSR-VTT (50.8\%), VATEX (64.5\%), MSVD (49.7\%), and DiDeMo (45.8\%).
%
The code is available at \href{https://github.com/bofang98/UATVR}{\texttt{\textcolor[rgb]{1,0,0.6}{https://github.com/bofang98/UATVR}}}.
\end{abstract}

\section{Introduction}
\label{sec:intro}

\footnote{
$^*$Co-first authorship. This work was done when Bo Fang was a
research intern at Baidu Inc.
}
\footnote{
$^{\dag}$Corresponding author.
}

\begin{figure}
    \centering
    \includegraphics[width=\linewidth]{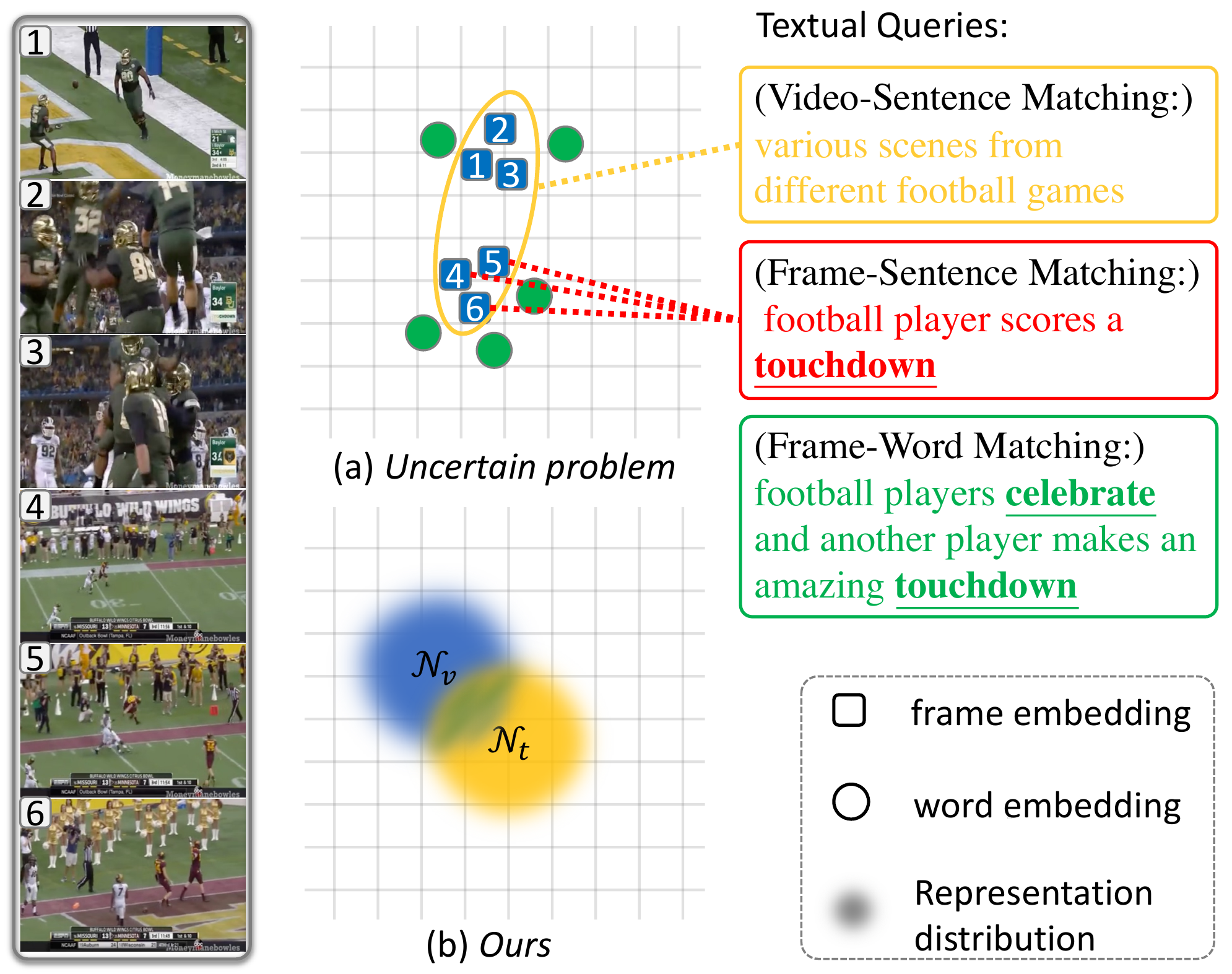}
    \caption{
    Motivation.
    A video has numerous descriptions containing different level information, (a) which shows inconsistent text-video correspondences in the common embedding space.
    The former three frames depict a `celebrate' action, while the last three frames are about a `touchdown'. This video diversity thus makes the optimal text-video matching in uncertain granularities, which we call an uncertain matching problem.
    Moreover, previous deterministic works can only handle one-to-one text-video mappings, yet a realistic relationship between two modalities is one-to-many.
    (b) The above problems motivate our uncertainty-adaptive model through distribution matching procedures.}
    \label{fig:overview}
\end{figure}

%
With surging portable filming devices and emerging video media platforms, searching videos of interest with human instructions, typically as texts, has been a part of daily lives, which urgently requires effective and robust text-video retrieval (TVR) techniques.
Given a query text (video), TVR aims to find the most relevant video (text) in the database, which is typically overwhelmed with sophisticated vague semantic combinations varying with hierarchical text structures or spatiotemporal video spans. 

%
Recent breakthroughs in the large-scale image and/or text pre-training~\cite{radford2021clip, jia2021ALIGN, yuan2021florence} benefit TVR significantly.
A serial of seminal works employ a separated encoder architecture to respectively project texts and videos into a pre-trained joint embedding space for compact cross-modal interaction \cite{lei2021clipbert,bain2021frozen,luo2021clip4clip, li2023tgvqa}. 

Since a video inherently contains information beyond texts, simply pooling all frames as a whole video expression brings distraction during matching specific text entities~\cite{luo2021clip4clip, gabeur2020mmt}.
Therefore, inspired by fine-grained image-text pre-training, \textit{e.g.}, FILIP~\cite{yao2021filip} and ALBEF~\cite{li2021ALBEF}, multi-grained TVR paradigms are introduced to build multi-level cross-modal interactions with \textit{sentence-frame} level~\cite{gorti2022xpool,lin2022textadaptive}, \textit{word-frame} level~\cite{wang2022disentangled}, or hierarchical correspondences including \textit{phrase-clip} level~\cite{min2022hunyuan_tvr, 
 jin2023DiCoSA}.
However, these methods are still far from satisfying in handling the intrinsic uncertainties of determining the optimal entity combinations with appropriate granularities during text-video matching.

In Fig.~\ref{fig:overview}, we illustrate the uncertain matching problem in TVR. Since different frame/word combinations can plausibly correspond to semantics in various perspectives, given the same video, successful retrieval can achieve
in varying granularities involving discrepant text-video entities, \ie, video-sentence matching, frame-sentence matching, frame- word matching, \textit{etc}.
%
%
%
Previous works determine particular cross-modal mapping strategies in certain granularities, yet none have studied the intrinsic uncertainties of optimal text-video entity combinations.
Besides, existing deterministic cross-modal retrieval can only handle one-to-one mapping scenarios~\cite{chun2021pcme}. However, a video can be described by multiple sentences typically (and vice versa), which formulates realistic one-to-many relationships.
%
%
%
%

In this paper, we propose a novel TVR framework to tackle the uncertainty problem in cross-modal matching, termed \textbf{U}ncertainty-\textbf{A}daptive \textbf{T}ext-\textbf{V}ideo \textbf{R}etrieval (\textbf{UATVR}).
Generally, UATVR models each text-video lookup as a distribution matching procedure in complementary deterministic and probabilistic views.
It is materialized upon word-frame token-wise interactions and consists of a dynamic semantic adaptation (DSA) module and a distribution-based uncertainty adaptation (DUA) module.

Concretely, DSA module enhances token-wise matching by introducing additional learnable multi-class tokens. 
We find these simple-yet-effective tokens can adaptively aggregate multi-grained video (or text) semantics during matching, thus allowing for flexible high-level reasoning.
For DUA, we represent samples from each modality as distributions rather than feature points and convert the deterministic matching process to probabilistic distribution alignment. 
To simulate one-to-many text-video mappings, we pull probabilistic embeddings sampled from each distribution closer via multi-instance contrastive loss~\cite{miech2020MIL-NCE}.

%

Our contributions can be summarized as  
(i) We innovatively model video and text representations as probabilistic distributions and align them through multiple-instance contrast in their common embedding space for uncertainty-adaptive cross-modal matching.
(ii) We propose a simple-yet-effective technique for flexible high-level reasoning by adding additional learnable tokens, allowing deterministic semantic uncertainty adaptation in videos/texts.
%
%
(iii) Comprehensive experimental explorations demonstrate the superiority of our UATVR, which obtains state-of-the-art results across public TVR benchmarks including MSR-VTT~\cite{xu2016msrvtt}, MSVD~\cite{wu2017msvd}, VATEX~\cite{wang2019vatex}, and DiDeMo~\cite{anne2017didemo}.

\section{Related Work}
\label{sec:relate}

\paragraph{Vision-Language Pre-training.}
%
Cross-modal vision language understanding~\cite{thomee2016yfcc100m,changpinyo2021conceptual} is a challenging task for both computer vision and natural language processing communities.
%
Recent breakthroughs are large-scale image-text contrastive pre-training, which employs a contrastive loss to jointly align image-text semantics into a unified embedding space, on more than 100M samples~\cite{radford2021clip,jia2021ALIGN}.
Vision-language pre-training with this paradigm~\cite{yuan2021florence, yu2022coca, li2021ALBEF, li2021declip, buch2022revisiting} has significantly boosted numerous cross-modal tasks such as VQA~\cite{antol2015vqa}, image captioning~\cite{xu2015imagecaption}, text-image retrieval~\cite{karpathy2015image-text-retrieval}, \textit{etc}.
For the video counterparts, large-scale video caption datasets, \textit{e.g.}, HowTo100M~\cite{miech2019howto100m} and WebVid2M~\cite{bain2021frozen}, also boost promising cross-modal video understanding.
However, due to the high cost of collecting wild videos and huge computing resources requirement, we bootstrap from CLIP like~\cite{luo2021clip4clip, portillo2021straightforward} for text-video retrieval.

\noindent \textbf{Text-Video Retrieval}
%
is to find the most semantic-relevant video given a text query (text $\rightarrow$ video).
Early research devotes to distilling knowledge from ``expert" models 
based on offline-extracted single-modality features~\cite{gabeur2020mmt, liu2019usewhathave, wang2021DMM, chen2020HGR,fang2022mamico, luo2022ERUV}.
%
Recent dominant TVR benefits from end-to-end pre-training on large-scale text-video datasets~\cite{miech2019howto100m, bain2021frozen, miech2020MIL-NCE}. 
Strategies that can improve training efficiency are essential for end-to-end paradigms like ClipBERT~\cite{lei2021clipbert} and Frozen~\cite{bain2021frozen}.
In TMVM~\cite{lin2022textadaptive}, masked-based prototypes for aggregating video features are proposed, which play a similar role to our DSA tokens. However, only visual RGB frames are modeled in TMVM, ignoring the hierarchical attributes in the textual counterpart.


Another idea of TVR transfers knowledge from publicly available CLIP models pre-trained on large-scale text-image pairs and then align text-video modalities with choreographed mapping strategies~\cite{luo2021clip4clip,zhao2022centerclip,gorti2022xpool, ma2022xclip,gao2021clip2tv,fang2021clip2video, chen2023tagging, jin2023videogameplayer, jin2023diffusionret}. 
%
%
Considering the discrepancy problem that videos always express more information than texts can capture~\cite{gorti2022xpool}, subsequent works devote to crafting cross-modal interactions with certain entities in specific granularities, \textit{e.g.}, sentence-frame level~\cite{gorti2022xpool,lin2022textadaptive}, word-frame level~\cite{wang2022disentangled}, and hierarchical level interactions~\cite{min2022hunyuan_tvr, jin2023DiCoSA, wu2023cap4video}.
TS2-Net~\cite{liu2022ts2net} selects top-$k$ informative tokens per frame, representing salient semantics, for frame-wise cross-modal matching. 
It is been designed upon a more fine-grained level.
Yet none above have studied the intrinsic uncertainties of optimal entity combinations in appropriate granularities, which motivates our uncertainty-adaptive matching model.

\noindent\textbf{Probabilistic Representations.}
The probabilistic theory has a long history in machine learning~\cite{murphy2012machinelearning}. 
For the vision domain, HIB~\cite{oh2018HIB} first introduces probabilistic embeddings to capture the uncertainty of image representations whilst handling the one-to-many correspondences for deep metric learning. 
Moreover, they have also been successfully applied to other tasks like face recognition~\cite{shi2019probabilistic-face, chang2020uncertain-facerecog}, pose estimation~\cite{sun2020prob-pose}, \etc.
%
PCME~\cite{chun2021pcme} employs probabilistic embeddings for text-image retrieval to perform one-to-many matching between the multiplicity of visual concepts, which inspires us to expand them to videos, as videos typically contain more complex semantic concepts for their temporal dynamics.
%
Moreover, we find that soft contrastive loss~\cite{oh2018HIB} used in PCME is sub-optimal for text-video modelling. 
Instead, we introduce multi-instance contrast for a more appropriate one-to-many relation simulating.
From this, our uncertainty-adaptive matching model tackles the uncertain matching problem and remarkably surpasses previous methods.

\section{Method}
\label{sec:method}

In this section, we first introduce our token-wise word-frame matching baseline. Then we propose two essential modules of UATVR, \eg, dynamic semantic adaptation and adaptive distribution matching, for tackling the uncertainty problem in text-video retrieval.

\begin{figure*}
    \centering
    \includegraphics[width=\linewidth]{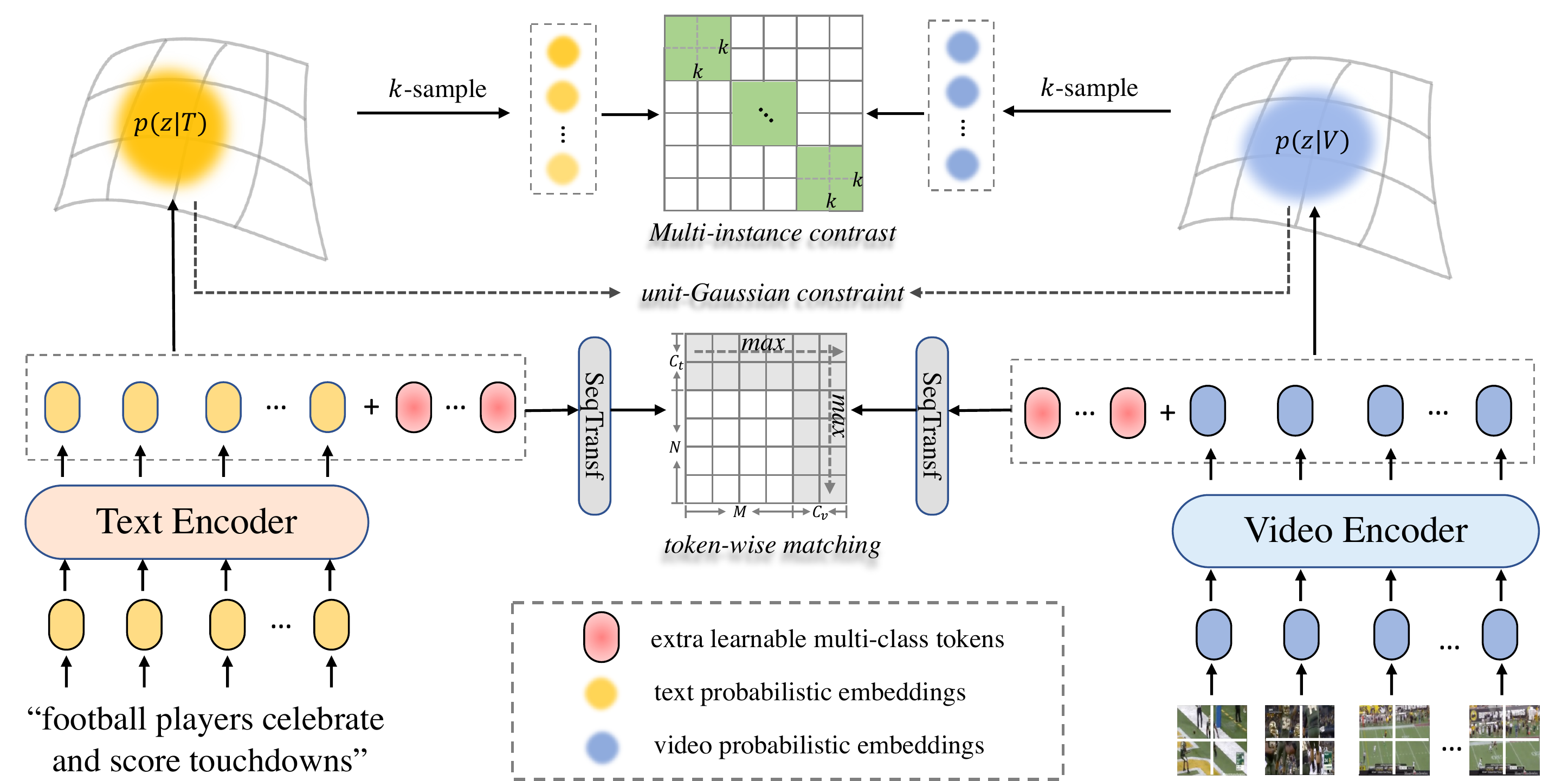}
    \caption{
    UATVR pipeline.
    We concatenate numerous extra learnable tokens with sequential frame/word embeddings and feed them into a lightweight SeqTransf~\cite{luo2021clip4clip} for adaptive token-wise matching.
    Besides, we model all visual and textual tokens as probabilistic distributions and sample $k$ probabilistic embeddings from each distribution with gradient propagating to construct multi-instance contrasts in a batch.
    }
    \label{fig:method}
\end{figure*}

\subsection{Preliminary}
\label{preliminary}

\textbf{Problem Definition.}
TVR aims to learn a similarity calculation function $s(\cdot)$, which ought to maximize the similarity score of positive cross-modal samples and assign lower similarity for irrelevant pairs.
Formally, given a pair of text $t_i\in \mathbb{R}^{N+1}$ and video $v_i\in \mathbb{R}^{M\times 3\times H\times W}$, we formulate them as collections of $N$ words and $M$ frames with $t_i=[w_i^0, w_i^1, w_i^2, \cdots, w_i^N]^T$, $v_i=[f_i^1, f_i^2, \cdots, f_i^M]^T$, where $w_i^0$ represents the \texttt{[CLS]} token and $H\times W$ denotes the resolution.
We feed $t_i$ and $v_i$ into a text encoder and a video encoder respectively to get their corresponding embeddings $\mathbf{t}_i=[\mathbf{w}_i^0, \mathbf{w}_i^1, \cdots, \mathbf{w}_i^N]^T$ and $\mathbf{v}_i=[\mathbf{f}_i^1, \mathbf{f}_i^2, \cdots, \mathbf{f}_i^M]^T$. 
The frame embedding $\mathbf{f}_i^m$ comes from the distinct \texttt{[CLS]} token from the transformer-based vision encoder for the \textit{m}th frame.
Normally, we represent the whole video by average pooling all frame embeddings in Eq.~\ref{eq:meanpool} and represent the sentence with the first \texttt{[CLS]} token feature $\mathbf{w}_i^0$ in Eq.~\ref{eq:word-cls}.
The similarity of the text-video is calculated as the inner production of $\mathbf{t}_i, \mathbf{v}_i$, \textit{c.f.} Eq.~\ref{eq:sim}.
\begin{equation}
    \mathbf{v}_i = \mathrm{meanPool}([\mathbf{f}_i^1, \mathbf{f}_i^2, \cdots, \mathbf{f}_i^M]^T),
    \label{eq:meanpool}
\end{equation}
\begin{equation}
    \mathbf{t}_i = \mathbf{w}_i^0,
    \label{eq:word-cls}
\end{equation}
\begin{equation}
    s(\mathbf{t}_i, \mathbf{v}_i) = \langle \mathbf{t}_i, \mathbf{v}_i \rangle.
    \label{eq:sim}
\end{equation}

In training, a common optimizing method is to use a symmetric cross-entropy loss in both text-to-video and video-to-text directions. 
Given a batch of $B$ text-video pairs, the model updates its parameters by maximizing the sum of the main diagonal of a $B\times B$ similarity matrix:
\begin{equation}
    \mathcal{L}_{t2v} = -\frac{1}{B}\sum_{i}^{B} \mathrm{log} \frac{\mathrm{exp}(s(\mathbf{t}_i, \mathbf{v}_i))}{\sum_{j=1}^{B} \mathrm{exp} (s(\mathbf{t}_i, \mathbf{v}_j))},
    \label{eq:t2v}
\end{equation}
\begin{equation}
    \mathcal{L}_{v2t} = -\frac{1}{B}\sum_{i}^{B} \mathrm{log} \frac{\mathrm{exp}(s(\mathbf{v}_i, \mathbf{t}_i))}{\sum_{j=1}^{B} \mathrm{exp} (s(\mathbf{v}_i, \mathbf{t}_j))}, 
    \label{eq:v2t}
\end{equation}
\begin{equation}
    \mathcal{L} = \mathcal{L}_{t2v} + \mathcal{L}_{v2t}.
    \label{eq:t2v-v2t}
\end{equation}

\textbf{Fine-grained Interactions.}
Before the late cross-modal fusion, the key point lies in how to extract accurate video information best described by corresponding textual queries.
The naive method pools all frame embeddings equally~\cite{luo2021clip4clip} yet ignores the discrepancy problem that a video contains more information than a single caption can depict~\cite{gorti2022xpool}. 
%
Recent work attempts to ameliorate the above problem by devoting to constructing multi-grained cross-modal interactions, including sentence-frame~\cite{gorti2022xpool,lin2022textadaptive}, word-frame~\cite{wang2022disentangled}, or multiple hierarchical interactions~\cite{min2022hunyuan_tvr,ma2022xclip}. 
In this paper, we take the token-wise word-frame matching paradigm as a solid baseline due to its successful application in image-text pre-training~\cite{yao2021filip} and strong TVR performance~\cite{wang2022disentangled}. 
The text-video similarity thus comes from the mean of the maximum similarities between each frame with all word-level embeddings in bi-directions, formulated as:
\begin{equation}
    s(\mathbf{t}_i, \mathbf{v}_i) = \frac{1}{2}
    \left(
    \sum_{n=1}^N \mathop{\mathrm{max}}\limits_{m=1}^M \langle \mathbf{w}_i^n, \mathbf{f}_i^m \rangle + 
    \sum_{m=1}^M \mathop{\mathrm{max}}\limits_{n=1}^N \langle \mathbf{w}_i^n, \mathbf{f}_i^m \rangle
    \right),
    \label{eq:token-wise-ti}
\end{equation}
where $M,N$ denote frame and word number in the $i$th sample pair. 
$\mathbf{w}_i^n, \mathbf{f}_i^m$, which are channel-wise normalized before calculating, refer to the $n$th word embedding and $m$th frame embedding, respectively.
Eq.\ref{eq:token-wise-ti} would produce larger similarity sums for longer video and text input. Therefore, an average operation is attached before addition.


\subsection{Dynamic Semantic Adaptation}
To a certain extent, vanila token-wise matching brings more accurate text-video correspondences. 
However, fine-grained TVR interaction in a deterministic matching granularity does not consider the uncertain matching problem.
%
To tackle the problem, 
we introduce multiple additional learnable tokens to dynamically aggregate multi-level video and text information while retaining the advance of local context matching in the token-wise baseline, \textit{c.f.} Fig.~\ref{fig:method}. 

Given sequential frame embeddings $\{\textbf{f}_i^m\}_{m=1}^M$ and word embeddings $\{\textbf{w}_i^n\}_{n=1}^N$ extracted from backbone encoders, we append additional $C_v$ and $C_t$ learnable multi-class tokens respectively at the sequences' beginning (or end).
The extra class tokens are randomly initialized and the position embeddings are omitted for brevity in Fig.~\ref{fig:method}.
Then we feed the enlarged $(M+C_v)$ frame embeddings set into another lightweight sequential Transformer (following the same seqTransf structure in~\cite{luo2021clip4clip}) to further model relations between learnable tokens and frames depending on the corresponding text queries.
We adopt the same operation symmetrically for the $(N+C_t)$ word embeddings.
All enlarged frame and word embeddings are channel-wise normalized before similarity calculation.
Similar to token-wise interactions, we finally calculate our modified text-video similarity function upon the union of frame/word tokens and extra learnable class tokens, formulated as:
\begin{equation}
    s = \frac{1}{2}
    \Bigg(
    \sum_{n=1}^{N+C_t} \mathop{\mathrm{max}}\limits_{m=1}^{M+C_v} \langle \mathbf{w}_i^n, \mathbf{f}_i^m \rangle 
    + 
    \sum_{m=1}^{M+C_v} \mathop{\mathrm{max}}\limits_{n=1}^{N+C_t} \langle \mathbf{w}_i^n, \mathbf{f}_i^m \rangle 
    \Bigg). \\
    \label{eq:DSA}
\end{equation}

We define our dynamic semantic adaption loss $\mathcal{L}_{\mathrm{DSA}}$ in the same formulation as Eq.~\ref{eq:t2v},\ref{eq:v2t},\ref{eq:t2v-v2t}, in which the modified similarity function in Eq.~\ref{eq:DSA} is employed for text-video cross-modal matching. 
The additional semantic-aggregated tokens introduce negligible parameters and training overhead, Tab.~\ref{tab:params}, which is entirely simple yet effective.
We give a deep analysis of DSA tokens in Sec.~\ref{ablation} and appendix, emphatically interpreting their working mechanisms.

\subsection{Adaptive Distribution Matching}
Since deterministic methods can only handle one-to-one mapping scenarios~\cite{chun2021pcme}, videos typically have multiple descriptions, formulating realistic one-to-many text-video relationships instead.
We propose adaptive distribution matching for probabilistic TVR to tackle the inherent inconsistency in text-video distributions, \cf Fig.~\ref{fig:method}.

Let $\mathbf{t}_i$, $\mathbf{v}_i$ denote the output of each backbone. We represent the text caption $t_i$ and the video $v_i$ as normal distributions $p(z|t_i)$ and $p(z|v_i)$ with mean vectors and diagonal covariance matrices in $\mathbb{R}^D$, respectively:
\begin{equation}
\begin{split}
    p(z|t_i) &\sim \mathcal{N} (h_{\mathcal{T}}^{\mu}(\mathbf{t}_i), 
                    \mathrm{diag}(h_{\mathcal{T}}^{\sigma}(\mathbf{t}_i))), \\
    p(z|v_i) &\sim \mathcal{N} (h_{\mathcal{V}}^{\mu}(\mathbf{v}_i), 
                    \mathrm{diag}(h_{\mathcal{V}}^{\sigma}(\mathbf{v}_i))),          
\end{split}
\label{eq:distribution}
\end{equation}
where head module $h^\mu$ is a fully-connected layer followed by LayerNorm~\cite{ba2016layernorm} and $l_2$ normalization, and head $h^\sigma$ is a separate fully-connected layer without any normalization following~\cite{chun2021pcme}. 
The textual head $h_{\mathcal{T}}$ and visual head $h_{\mathcal{V}}$ share the same parametric structure but optimize independently upon transformer-based feature output. Next, two groups of $K$ probabilistic embeddings 
$\{\mathbf{t}_i^{(1)}, \cdots, \mathbf{t}_i^{(K)} \}\overset{\mathrm{iid}}{\sim} p(z|t_i)$ and 
$\{\mathbf{v}_i^{(1)}, \cdots, \mathbf{v}_i^{(K)} \}\overset{\mathrm{iid}}{\sim} p(z|v_i)$ are generated by sampling from the distributions of $t_i$ and $v_i$ with gradient propagating.
To enable stable training, we use the reparametrization trick~\cite{kingma2013auto} during the generation, formulated as follows:
\begin{equation}
\begin{split}
    \mathbf{t}_i^{(k)} &= \sigma (t_i)\cdot \epsilon ^{k} + \mu (t_i), \\
    \mathbf{v}_i^{(k)} &= \sigma (v_i)\cdot \epsilon ^{k} + \mu (v_i),
\end{split}
\label{eq:sample}
\end{equation}
where $\epsilon ^{(k)} \overset{\mathrm{iid}}{\sim} \mathcal{N}(0, I)$ and $\mu, \sigma$ denote the mean and the standard deviation of $p(z|t_i)$, $p(z|v_i)$.

Unlike previous methods~\cite{oh2018HIB,chun2021pcme} using soft contrastive loss (a binary classification loss based on the softmax cross-entropy) via Monte-Carlo estimation for distribution alignment, we treat all probabilistic embeddings from a matched text-video pair as positive samples to simulate one-to-many cross-modal matching. 
We update the model with a Multi-Instance InfoNCE~\cite{miech2020MIL-NCE} loss, and the training target is to minimize the distance between video (text) probabilistic embeddings and all corresponding text (video) embeddings. 
Further comparisons are made in the appendix.
Given a specific textual probabilistic embedding $\mathbf{t}_i\in \{\mathbf{t}_i^{(k)}\}_{k=1}^K$, we define the positive set $\mathcal{P}_i$ for $\mathbf{t}_i$ as all video probabilistic embeddings from $v_i$, formulated as $\mathcal{P}_i = \{\mathbf{v}_i^{(k)}\}_{k=1}^K $.
The negative set thus is formed as probabilistic embeddings from other videos in the batch, $\widetilde{\mathcal{P}}_i = \{\mathbf{v}_j^{(k)}\}_{j,k}, j\neq i$.
We define the distribution-based uncertainty adaptation loss as:
\begin{equation}
    \mathcal{L}_{\mathrm{DUA}} = 
    -\frac{1}{B}\sum_{i}^{B}
    \mathrm{log}
    \frac{\sum_{\mathbf{v}_i \in \mathcal{P}_i} \mathrm{exp} (s(\mathbf{t}_i, \mathbf{v}_i))}
    {\sum_{\mathbf{v}_j \in \{ \mathcal{P}_i \cup \widetilde{\mathcal{P}}_i \} } \mathrm{exp} (s(\mathbf{t}_i, \mathbf{v}_j))}.
\end{equation}


\subsection{Total Objectives}
Following~\cite{oh2018HIB}, we introduce additional KL divergence loss between the distributions and the unit Gaussian prior $\mathcal{N}(0, I)$ to constraint the learned variances from collapsing to zero, which can be formulated as:
\begin{equation}
    \mathcal{L}_{\mathrm{KL}} = \mathrm{KL}(p(z|t_i), \mathcal{N}(0, I)) + \mathrm{KL}(p(z|v_i), \mathcal{N}(0, I)).
    \label{eq:KL}
\end{equation}
Therefore, the total objectives can be defined as:
\begin{equation}
    \mathcal{L} = \mathcal{L}_{\mathrm{DSA}} + 
    \alpha \cdot \mathcal{L}_{\mathrm{DUA}} + 
    \beta \cdot \mathcal{L}_{\mathrm{KL}},
    \label{eq:all}
\end{equation}
where $\alpha$ and $\beta$ control the trade-off among three terms.


\begin{table}[t]
  \begin{center}
  \setlength\tabcolsep{4pt}
  \begin{tabular}{l|ccccc}
    \Xhline{0.7pt}
      Methods  &
      R@1 & 
      R@5 & 
      R@10 & 
      MdR$\textcolor{green}{\downarrow}$ & 
      MnR$\textcolor{green}{\downarrow}$  \\
    \hline
    CLIP4Clip~\cite{luo2021clip4clip}  & 47.1 & 74.1 & 81.8 & 2.0 & 14.9 \\
    TI (Token-Wise)  & 48.4 & 74.2 & 83.3 & 2.0 & 14.1 \\
    \hline
    + DSA  & 49.6 & 75.5 & 84.9 & 2.0 & 12.5 \\ 

    + DUA$^{\dag}$  & 50.1 & 75.8 & 84.6 & 1.5 & 12.8  \\
    \rowcolor{mygray} + KL$^{\dag}$ (UATVR) & \textbf{50.8} & \textbf{76.3} & \textbf{85.5} & \textbf{1.0} & \textbf{12.4} \\
    \hline
    + DUA*  & 50.0 & 75.8 & 83.9 & 1.5 & 12.9  \\
    + KL*  & 50.6 & 75.9 & 84.9 & \textbf{1.0} & 12.8 \\
    \Xhline{0.7pt}
  \end{tabular}
  \end{center}
  \caption{
  Ablation study of different components.
  $^{\dag}$ denotes the implementation with MIL-NCE contrast and
  * is implemented with soft contrastive loss via Monte-Carlo estimation~\cite{oh2018HIB}.}
  \label{tab:ablation-prob-loss}
\end{table}
\section{Experiments}
\label{sec:exps}
We first describe the experimental settings. Then thorough ablation studies are conducted to demonstrate the effectiveness of our proposed UATVR. Finally, we make comparisons of our model to existing state-of-the-art methods on various TVR benchmarks.

\subsection{Experimental Settings}
\label{setting}

\noindent \textbf{Datasets.} 
Experiments are conducted on 4 common video-text retrieval benchmarks:
\textbf{(a) MSR-VTT}~\cite{xu2016msrvtt} contains 10K video clips in total with 20 captions for each. Following the data splits from ~\cite{gabeur2020mmt, miech2019howto100m, luo2021clip4clip}, we train models on the \texttt{Training-9K} set with corresponding captions and report results on the \texttt{test 1K-A} set.   
\textbf{(b) MSVD}~\cite{wu2017msvd} includes 1,970 videos and 80K captions, with $\sim$40 captions on average per video. Train, validation, and test set have 1,200, 100, and 670 videos respectively.
\textbf{(c) DiDeMo}~\cite{anne2017didemo} contains 10K videos paired with 40K descriptions. Following previous~\cite{luo2021clip4clip, bain2021frozen, lei2021clipbert}, we concatenate all descriptions of one video to a single query.
\textbf{(d) VATEX}~\cite{wang2019vatex} collects $\sim$35K videos with multiple annotations for each. There are $\sim$26K videos for training, 1,500 for validating, and 1,500 for testing.

\noindent \textbf{Evaluation Metrics.} 
For brevity, we abbreviate Recall at $K$ to R@$K$ ($K=1,5,10$) upon all datasets, which calculates the percentage of correct videos among the top $K$ retrieved videos given textual queries (Text$\rightarrow$Video, and vice versa). 
MdR, Median Rank, calculates the median of the ground truth in the retrieval ranking list. 
MnR, Mean Rank, calculates the mean rank of the correct results in the retrieval ranking list. 
Note that for MdR and MnR, the lower score means the better (indicated as $\textcolor{green}{\downarrow}$).

\begin{table}[t]
    \begin{center}
     \setlength\tabcolsep{2pt}   
    \begin{tabular}{l|c|c|c}
    \Xhline{0.7pt}
    Methods & Time Complexity & Params & Time\\
    \hline
       CLIP4Clip~\cite{luo2021clip4clip} & $\mathcal{O}(B^{2})$  & 162.3M & 75.04h \\
       TI (Token-Wise) & $\mathcal{O}(B^{2}MN)$ & 162.8M & 82.00h \\
       w/ DSA & $\mathcal{O}(B^{2}M'N')$ & 162.8M & 82.24h \\
       w/ DUA & $\mathcal{O}(B^2(M'N'+K^{2}))$ & 164.9M & 84.08h \\
    \Xhline{0.7pt}
    \end{tabular}
    \end{center}
    \caption{
    Comparisons of different components. 
    $B$ denotes sample size.
    $M, N$ denote the length of frame tokens and text tokens resp.
    $M', N'$ are slightly enlarged with additional learnable tokens.
    $K$ is the number of probabilistic embeddings. 
    Training time denotes GPU hours calculated by a single P40 card.
    }
    \label{tab:params}
    \vspace{-5mm}
\end{table}

        

\noindent \textbf{Implementation Details.} 
We initialize our visual and language backbone with CLIP~\cite{radford2021clip} pre-trained weight.
Following ~\cite{luo2021clip4clip}, we further use a four-layer lightweight sequential transformer to encode extra learnable class tokens with frame and word embeddings. 
In ablations, we take ViT/B-16 by default.
The textual token length is 32 and the frame length is 12 for all datasets except DiDeMo (64 max query words and 64 frames). 
A uniform frame sampling strategy with one frame per second sampling rate is employed.
The dimension of video (text) distributions is 512 by default.
Following~\cite{luo2021clip4clip,gorti2022xpool,liu2022ts2net}, we train UATVR model for 5 epochs with Adam~\cite{kingma2015adam} optimizer and adopt a warmup~\cite{goyal2017warmup} strategy.
We set the batch size as 64 and the initial learning rate as 5e-5.
The coefficient is 1e-2 for $\alpha$ and 1e-4 for $\beta$.


\begin{figure}[t]
    \centering
    \includegraphics[width=\linewidth]{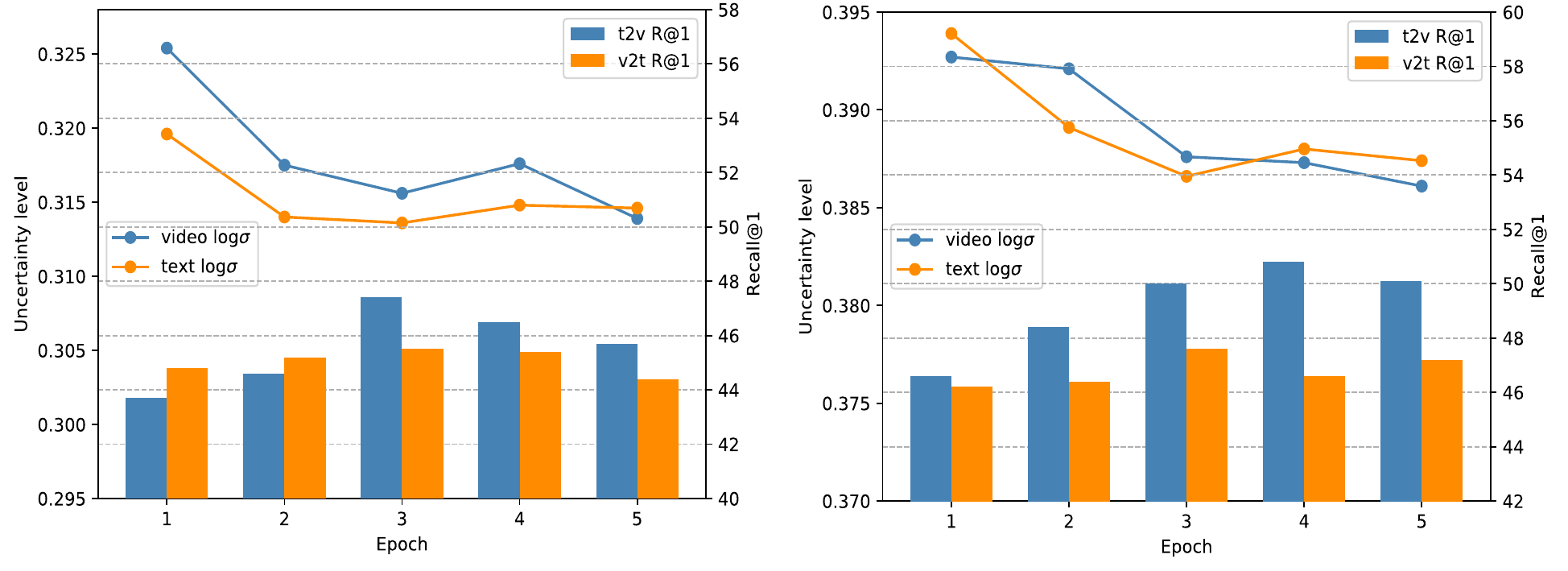}
    \caption{Uncertainty level versus R@1 on MSR-VTT dataset. (Left) Results on ViT-B/32. (Right) Results on ViT-B/16. }
    \label{fig:distribution}
\end{figure}

\begin{table}[t]
  \begin{center}    
  \setlength\tabcolsep{4pt}
    \begin{tabular}{cc|ccccc}
    \Xhline{0.7pt}
    \multicolumn{2}{c}{Extra-Tokens \#}  \vline& R@1 & R@5 & R@10 & MdR$\textcolor{green}{\downarrow}$ & MnR$\textcolor{green}{\downarrow}$ \\
    \hline
    baseline & \textcolor{gray}{0}   & 48.4 & 74.2 & 83.3 & 2.0 & 14.1 \\
    \hline
    & 1 & 48.9 & 75.3 & 84.6 & 2.0 & 12.3 \\
    & 2 & \textbf{49.6} & 75.5 & 84.6 & 2.0 & 12.5 \\
    \rowcolor{mygray}
    $C_v$ &  3 & \textbf{49.6} & \textbf{75.7} & 84.6 & 2.0 & \textbf{11.9} \\
    $(C_t=0)$& 4 & 49.3 & 75.2 & 84.2 & 2.0 & 12.4 \\
    & 8 & 48.8 & 75.2 & 84.7 & 2.0 & 12.5 \\
    & 12 & 49.2 & 75.4 & 84.2 & 2.0 & 12.4 \\
    \hline
    \rowcolor{mygray}
    $C_t$ &  2 & \textbf{49.6} & 75.5 & \textbf{84.9} & 2.0 & 12.5 \\
    $(C_v=3)$& 4 & 49.1 & 75.4 & 84.4 & 2.0 & 13.3\\
    \Xhline{0.7pt}
  \end{tabular}
  \end{center}
  \caption{Ablation study for the number of extra learnable tokens.}
  \label{tab:ablation-cls-num}
  \vspace{-3mm}
\end{table}

\subsection{Ablation Study}
\label{ablation}
We evaluate the effectiveness of different components in UATVR by comprehensive experiments.
The default visual encoder is ViT-B/16~\cite{dosovitskiy2020ViT}
and the \textit{t}2\textit{v} retrieval results are reported on the widely-used MSR-VTT~\cite{xu2016msrvtt} dataset.

\noindent \textbf{Uncertainty-Adaptive Matching.}
The baseline of UATVR is fine-grained token-wise interaction (TI), which provides minimal granularity tokens for retrieval. 
%
%
As shown in Tab.~\ref{tab:ablation-prob-loss}, with extra multi-class learnable tokens appended, we observe a 1.2\% R@1 improvement (48.4\% \textit{vs.} 49.6\%) and a lower MnR (14.1 \textit{vs.} 12.5) compared to the baseline. We explain that additional tokens can aggregate multi-grained information extracted from video frames and textual words, which adapts to flexible cross-modal matching in different granularities.
Moreover, distribution-based uncertainty adaptation with KL divergence constraint further obtains the highest 50.8\% R@1 and the lowest 1.0 MdR and 12.4 MnR, which demonstrates the effectiveness of the proposed distribution alignment mechanism.
%
Also, we formulate a soft contrastive loss following~\cite{oh2018HIB}, which is a binary classification loss based on the softmax cross-entropy via Monte-Carlo estimation (marked as *). 
We observe very close but slightly lower performance than MIL-NCE (multi-instance contrast, marked as †).
Further analysis is shown in the appendix.
All the above results prove the significance of distribution-based cross-modal matching in tackling the proposed uncertainty problem.

Tab.~\ref{tab:params} compares time complexity and params for each component, in which feature dimension $D$ is omitted for brevity.
%
Despite the relatively larger time complexity taken by DSA and DUA, it is still limited in quadratic time for one text-video pair.
Note that our extra learnable tokens do not bring more model parameters, meanwhile having negligible additional training cost.
A similar conclusion can be drawn for the distribution-based matching module.
Therefore, our proposed UATVR is simple-yet-effective.

\begin{table}[t]
  \begin{center}
  \setlength\tabcolsep{3.5pt}
    \begin{tabular}{rc|ccccc}
    \Xhline{0.7pt}
    \multicolumn{2}{c}{Prob-Embeds \#}  \vline& R@1 & R@5 & R@10 & MdR$\textcolor{green}{\downarrow}$ & MnR$\textcolor{green}{\downarrow}$ \\
    \hline
    \multicolumn{2}{c}{($C_v=3,C_t=2$)}   \vline& 49.6 & 75.5 & 84.9 & 2.0 & 12.5 \\
    \hline
    \multirow{5}{*}{$K$} 
    & 1 & 49.6 & 76.5 & 84.3 & 2.0 & 12.5 \\
    & 3 & 49.8 & 76.1 & 84.9 & 2.0 & 12.9 \\
    & 5 & 50.5 & \textbf{77.1} & 84.4 & \textbf{1.0} & 12.6 \\
    \rowcolor{mygray} 
    & 7 & \textbf{50.8} & 76.3 & \textbf{85.5} & \textbf{1.0} & \textbf{12.4 }\\
    & 9 & - & - & - & - & - \\
    \Xhline{0.7pt}
  \end{tabular}
  \end{center}
  \caption{Ablation for the number of probabilistic embeddings.}
  \label{tab:ablation-prob-num}
\end{table}

\begin{table}[t]
  \begin{center}
    \begin{tabular}{c|ccccc}
    \Xhline{0.7pt}
    Frames \# & R@1 & R@5 & R@10 & MdR$\textcolor{green}{\downarrow}$ & MnR$\textcolor{green}{\downarrow}$ \\
    \hline
    4 & 44.9 & 74.0 & 82.2 & 2.0 & 15.1 \\
    8 & 50.6 & 76.0 & 83.9 & \textbf{1.0} & 12.9 \\
    \rowcolor{mygray} 12 & 50.8 & 76.3 & \textbf{85.5} & \textbf{1.0} & \textbf{12.4} \\
    16 & \textbf{51.0} & \textbf{76.4} & \textbf{85.5} & \textbf{1.0} & 13.5 \\
    20 & 50.8 & 76.0 & 84.9 & \textbf{1.0} & 13.4 \\
    \Xhline{0.7pt}
  \end{tabular}
  \end{center}
  \caption{Impact of visual frame numbers.}
  \label{tab:ablation-frame-num}
\vspace{-5mm}
\end{table}

\noindent \textbf{Dynamic Semantic Adaptation Tokens.}
In Tab.~\ref{tab:ablation-cls-num}, we study the impact of additional appended $C_v$ and $C_t$ multi-class learnable tokens.
It shows distinct \textit{t}2\textit{v} retrieval improvements once extra visual tokens are added (\textit{i.e.}, $C_v>0$), which reflects that this simple-yet-effective technology can aggregate multi-grained video semantics depending on the uncertain captions.
$C_v=3$ is the best among all. 
When $C_v$ is larger than 3, the performance starts to degrade.
We observe a similar phenomenon in TMVM~\cite{lin2022textadaptive} that more video prototypes would significantly degrade the performance. 
Too many extra tokens would become noise rather than representative, negatively influencing the  normal matching process.
In subsequent experiments, we set final $C_v=3$ and $C_t=2$. 
Additional $C_t$ text tokens further promote \textit{v}2\textit{t} R@1 to 47.8\%. 
We analyze the corresponding \textit{v}2\textit{t} retrieval results in the appendix.


\begin{table*}[t]
  \begin{center}
    \setlength\tabcolsep{5pt}
    \begin{tabular}{lc|ccccc|ccccc}
    \Xhline{0.7pt}
    \multirow{2}{*}{Method} & \multirow{2}{*}{Date} & \multicolumn{5}{c}{Text $\rightarrow$ Video}   \vline& \multicolumn{5}{c}{Video $\rightarrow$ Text} \\
       & &  R@1 & R@5 & R@10 & MdR$\textcolor{green}{\downarrow}$ & MnR$\textcolor{green}{\downarrow}$  & R@1 & R@5 & R@10 & MdR$\textcolor{green}{\downarrow}$ & MnR$\textcolor{green}{\downarrow}$ \\
    \hline
    MMT~\cite{gabeur2020mmt} & ECCV'20 & 26.6 & 57.1 & 69.6 & 4.0 & - & 27.0 & 57.5 & 69.7 & 3.7 & 21.3 \\
    SupportSet~\cite{patrick2020supportset} & ICLR'21 & 30.1 & 58.5 & 69.3 & 3.0 & - & 28.5 & 58.6 & 71.6 & 3.0 & -\\
    Frozen~\cite{bain2021frozen} &  ICCV'21 & 32.5 & 61.5 & 71.2 & 3.0 & - & - & - & - & - & - \\
    BridgeFormer~\cite{ge2022bridgeformer} & CVPR'22 & 37.6 & 64.8 & 75.1 & - & - & - & - & - & - & - \\
    TMVM~\cite{lin2022textadaptive} & NeurIPS'22 & 36.2 & 64.2 & 75.7 & 3.0 & - & 34.8 & 63.8 & 73.7 & 3.0 & - \\
    \hline
    \textit{CLIP-ViT-B/32} & & & & & & & & & & & \\
    \hdashline[0.5pt/5pt]
    CLIP4Clip~\cite{luo2021clip4clip} & ArXiv'21 & 44.5 & 71.4 & 81.6 & 2.0 & 15.3 & 42.7 & 70.9 & 80.6 & 2.0 & 11.6 \\
    CenterCLIP~\cite{zhao2022centerclip} & SIGIR'22 & 44.2 & 71.6 & 82.1 & 2.0 & 15.1 & 42.8 & 71.7 & 82.2 & 2.0 & 10.9 \\
    CAMoE~\cite{cheng202camoe}$^{\ddagger}$ & ArXiv'21 & 44.6 & 72.6 & 81.8 & 2.0 & 13.3 & 45.1 & 72.4 & 83.1 & 2.0 & 10.0 \\
    CLIP2Video~\cite{fang2021clip2video} & ArXiv'21 & 45.6 & 72.6 & 81.7 & 2.0 & 14.6 & 43.5 & 72.3 & 82.1 & 2.0 & 10.2 \\
    X-Pool~\cite{gorti2022xpool} & CVPR'22 & 46.9 & 72.8 & 82.2 & 2.0 & 14.3 & - & - & - & - & - \\
    QB-Norm~\cite{bogolin2022qbnorm}$^{\ddagger}$ & CVPR'22 & 47.2 & 73.0 & 83.0 & 2.0 & - & - & - & - & - & - \\
    EMCL~\cite{jin2022EMCL} & NeurIPS'22 & 46.8 & 73.1 & 83.1 & 2.0 & - & 46.5 & 73.5 & 83.5 & 2.0 & -\\
    TS2-Net~\cite{liu2022ts2net} & ECCV'22 & 47.0 & \textbf{74.5 }& \textbf{83.8} & 2.0 & 13.0 & 45.3 & \textbf{74.1} & 83.7 & 2.0 & 9.2 \\
    \rowcolor{mygray} \textbf{UATVR} & & \textbf{47.5} & 73.9 & 83.5 & 2.0 & \textbf{12.3} & \textbf{46.9} & 73.8 & \textbf{83.8} & 2.0 & \textbf{8.6} \\
    \rowcolor{mygray} \textbf{UATVR}$^{\ddagger}$ &  & \textbf{49.8} & \textbf{76.1} & \textbf{85.5} & 2.0 & \textbf{12.9} & \textbf{51.1} & \textbf{74.8} & \textbf{85.1} & \textbf{1.0} & \textbf{8.3} \\
    \hline
    \textit{CLIP-ViT-B/16} & & & & & & & & & & & \\
    \hdashline[0.5pt/5pt]
    CLIP2TV~\cite{gao2021clip2tv} & ArXiv'21 & 48.3 & 74.6 & 82.8 & 2.0 & 14.9 & 46.5 & 75.4 & 84.9 & 2.0 & 10.2 \\
    CenterCLIP~\cite{zhao2022centerclip} & SIGIR'22 & 48.4 & 73.8 & 82.0 & 2.0 & 13.8 & 47.7 & 75.0 & 83.3 & 2.0 & 10.2 \\
    TS2-Net~\cite{liu2022ts2net} & ECCV'22 & 49.4 & 75.6 & 85.3 & 2.0 & 13.5 & 46.6 & 75.9 & 84.9 & 2.0 & 8.9 \\
    \rowcolor{mygray} \textbf{UATVR}  &  & \textbf{50.8} & \textbf{76.3} &  \textbf{85.5} & \textbf{1.0} & \textbf{12.4} & \textbf{48.1} & \textbf{76.3} & \textbf{85.4} & 2.0 & \textbf{8.0} \\
    \rowcolor{mygray} \textbf{UATVR}$^{\ddagger}$ &  & \textbf{53.5} & \textbf{79.5} & \textbf{88.1} & \textbf{1.0} & \textbf{10.2} & \textbf{54.5} & \textbf{79.1} & \textbf{87.9} & \textbf{1.0} & \textbf{7.6} \\
    \Xhline{0.7pt}
  \end{tabular}
  \end{center}
  \caption{\textit{t}2\textit{v} and \textit{v}2\textit{t} comparisons on MSR-VTT~\cite{xu2016msrvtt}. $^{\ddagger}$ denotes using inverted dual softmax~\cite{cheng202camoe} or QB-Norm~\cite{bogolin2022qbnorm} for post-processing.}
  \label{tab:sota-MSRVTT}
  \vspace{-3mm}
\end{table*}

\begin{table}
  \begin{center}
  \setlength\tabcolsep{3.5pt}
    \begin{tabular}{l|ccccc}
    \Xhline{0.7pt}
        Method & R@1 & R@5 & R@10 & MdR$\textcolor{green}{\downarrow}$ & MnR$\textcolor{green}{\downarrow}$ \\
        \hline
        HGR~\cite{chen2020HGR} & 35.1 & 73.5 & 83.5 & 2.0 & - \\
        CLIP~\cite{radford2021clip} & 39.7 & 72.3 & 82.2 & 2.0 & 12.8 \\
        SUPPORT~\cite{patrick2020supportset} & 44.9 & 82.1 & 89.7 & 1.0 & - \\
        CLIP4Clip~\cite{luo2021clip4clip} & 55.9 & 89.2 & 95.0 & 1.0 & 3.9 \\
        Clip2Video~\cite{fang2021clip2video} & 57.3 & 90.0 & 95.5 & 1.0 & 3.6 \\
        QB-Norm~\cite{bogolin2022qbnorm} & 58.8 & 88.3 & 93.8 & 1.0 & - \\
        TS2-Net~\cite{liu2022ts2net} & 59.1 & 90.0 & 95.2 & 1.0 & 3.5 \\
        \hline
        \rowcolor{mygray} UATVR(ViT-B32) & 61.3 & 91.0 & 95.6 & 1.0 & 3.3 \\
        \rowcolor{mygray} UATVR(ViT-B16) & \textbf{64.5} & \textbf{92.6} & \textbf{96.8} & 1.0 & \textbf{2.8} \\
    \Xhline{0.7pt}
  \end{tabular}
  \end{center}
  \caption{\textit{t}2\textit{v} comparisons on the \textbf{VATEX}~\cite{wang2019vatex} dataset.}
  \label{tab:sota-vatex}
  \vspace{-2mm}
\end{table}

\begin{table}
  \begin{center}
    \setlength\tabcolsep{3.5pt}
    \begin{tabular}{l|ccccc}
    \Xhline{0.7pt}
    Method & R@1 & R@5 & R@10 & MdR$\textcolor{green}{\downarrow}$ & MnR$\textcolor{green}{\downarrow}$ \\
    \hline
    ClipBERT~\cite{lei2021clipbert} & 20.4 & 48.0 & 60.8 & 6.0 & - \\
    TT-CE~\cite{croitoru2021teachtext} & 21.6 & 48.6 & 62.9 & 6.0 & - \\
    Frozen~\cite{bain2021frozen} & 31.0 & 59.8 & 72.4 & 3.0 & - \\
    TMVM~\cite{lin2022textadaptive} & 36.5 & 64.9 & 75.4 & 3.0 & - \\
    CLIP4Clip~\cite{luo2021clip4clip} & 42.8 & 68.5 & 79.2 & 2.0 & 18.9 \\
    TS2-Net~\cite{liu2022ts2net} & 41.8 & 71.6 & 82.0 & 2.0 & 14.8 \\
    \hline
    \rowcolor{mygray} UATVR(ViT-B32) & 43.1 & 71.8 & 82.3 & 2.0 & 15.1 \\
    \rowcolor{mygray} UATVR(ViT-B16) & \textbf{45.8} & \textbf{73.7} & \textbf{83.3} & 2.0 & \textbf{13.5} \\
    \Xhline{0.7pt}
  \end{tabular}
  \end{center}
  \caption{\textit{t}2\textit{v} comparisons on the \textbf{DiDeMo}~\cite{anne2017didemo} dataset.}
  \label{tab:sota-didemo}
  \vspace{-5mm}
\end{table}

\noindent \textbf{Distribution-Based Uncertainty Adaptation.}
A larger number of probabilistic embeddings can better simulate video and caption distributions but can also lead to more computing requirements.
In Tab.~\ref{tab:ablation-prob-num}, we report the performance according to the number of sampled probabilistic embeddings $K$ based on our optimal DSA branch.
We find that \textit{t}2\textit{v} retrieval performance increases as $K$ increases. 
When $K$ is larger than 7, the performance starts to saturate.
Considering the computational costs, we choose $K=7$ finally.
Moreover, we have attempted sampling different $K$ for text and video distributions in the appendix and obtain a similar conclusion to Tab.~\ref{tab:ablation-prob-num}.
Overall, our DUA surpasses the baseline by a large margin, which demonstrates the effectiveness of the distribution alignment mechanism.

In Fig.~\ref{fig:distribution}, we measure the inherent uncertainty (geometric mean over the $\sigma \in \mathbb{R}^{D}$) of test set texts and videos and report the R@1 performance in each epoch.
We show comparisons on two visual encoders to analyze the correlation between the uncertainty and the discriminability of learned representations.
Generally, we observe performance improvements with decreasing uncertainty, which verifies the positive effects of distribution-based uncertain adaptation.

\noindent \textbf{The Number of Visual Frames.}
The impact of frame numbers is studied in Tab.~\ref{tab:ablation-frame-num}. 
UATVR achieves a decent 50.6\% R@1 with only 8 frames.
The performance starts to saturate with more than 12 frames. 
Here, we only use 12 frames by default for fair comparisons with others.




\subsection{Comparison with State-of-the-arts}
\label{sota}
To evaluate the generalization of our uncertainty-adaptive models, we compare UATVR with SOTA methods on various text-video retrieval benchmarks, including MSR-VTT~\cite{xu2016msrvtt}, MSVD~\cite{wu2017msvd}, VATEX~\cite{wang2019vatex}, and DiDeMo~\cite{anne2017didemo}.

Tab.~\ref{tab:sota-MSRVTT} shows detailed comparisons on MSR-VTT \texttt{test 1k-A} set. 
We divide current approaches into Training-from-scratch (upper rows) and CLIP-Driven~\cite{radford2021clip}. 
Transferring knowledge from CLIP has distinctly surpassed models w/o initialization, which demonstrates that spatial semantics learned from image-text pairs are essential for the TVR task.
Our proposed UATVR falls into the CLIP-Driven paradigm.
For the ViT-B/32 encoder, UATVR obtains higher R@1 performance than the previous best method (47.5\% vs. 47.2\% in \textit{t}2\textit{v} retrieval, and 46.9\% vs. 46.5\% in \textit{v}2\textit{t} retrieval).
The improvement is more significant on the ViT-B/16 backbone.
Specifically, UATVR outperforms previous best TS2-Net~\cite{liu2022ts2net} by 1.4\% in \textit{t}2\textit{v} and 1.5\% in \textit{v}2\textit{t} retrieval, yielding a remarkable \textit{t}2\textit{v} R@1 50.8\%.
Notice that our method firstly reduces the MdR metric from 2.0 to 1.0 and has the lowest 12.4 MnR, which means UATVR is more robust to wrong retrieval samples.
In the appendix, we further report results with dual softmax learning (DSL) operation. 
Our results again surpass methods with post-processing operations like QB-Norm~\cite{bogolin2022qbnorm} and CAMoE~\cite{cheng202camoe}.

\begin{figure}
    \centering
    \includegraphics[width=\linewidth]{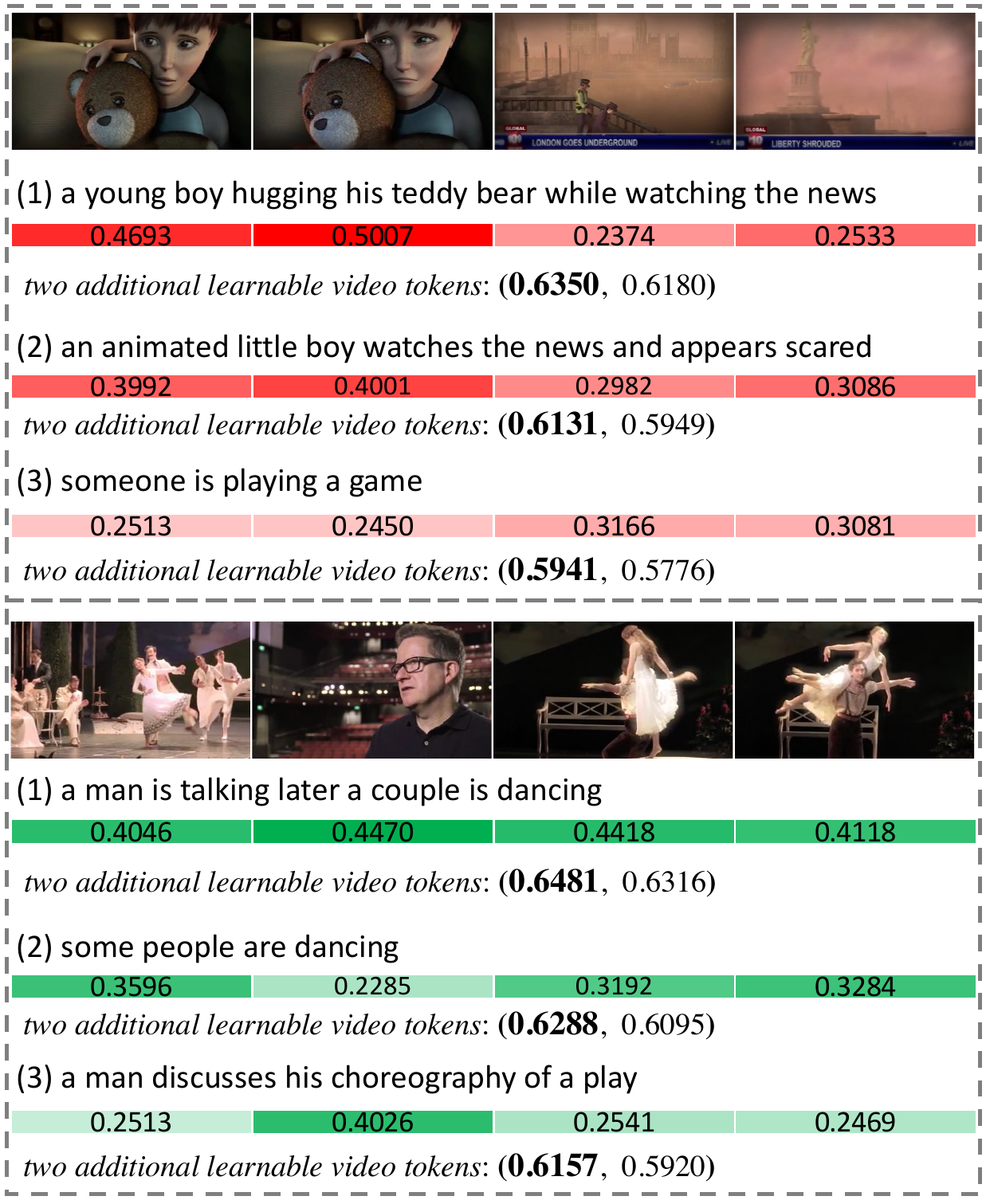}
    \caption{Impact of additional learnable tokens.
    Darker background color denotes higher frame attention.
    }
    \label{fig:extra-token-vis}
    \vspace{-5mm}
\end{figure}

Moreover, we conduct evaluations on multiple other TVR benchmarks, including VATEX~\cite{wang2019vatex} in Tab.~\ref{tab:sota-vatex}, DiDeMo~\cite{anne2017didemo} in Tab.~\ref{tab:sota-didemo}, and MSVD~\cite{wu2017msvd} in Tab.~\ref{tab:sota-msvd}.
Despite possible sub-optimal hyper-parameters (e.g., $C_v, C_t, K$) for the specific dataset, UATVR achieves consistent improvements across various datasets, \textit{e.g.}, 61.3\% \textit{vs.} 59.1\% for VATEX, and 43.1\% \textit{vs.} 42.8\% for DiDeMo. UATVR outperforms SOTAs by a large margin with better-performed ViT-B/16.
Prominent results demonstrate good generalization and robustness of our dynamic semantic-aggregation and distribution-based uncertainty adaption paradigms.

\begin{figure}
    \centering
    \includegraphics[width=\linewidth]{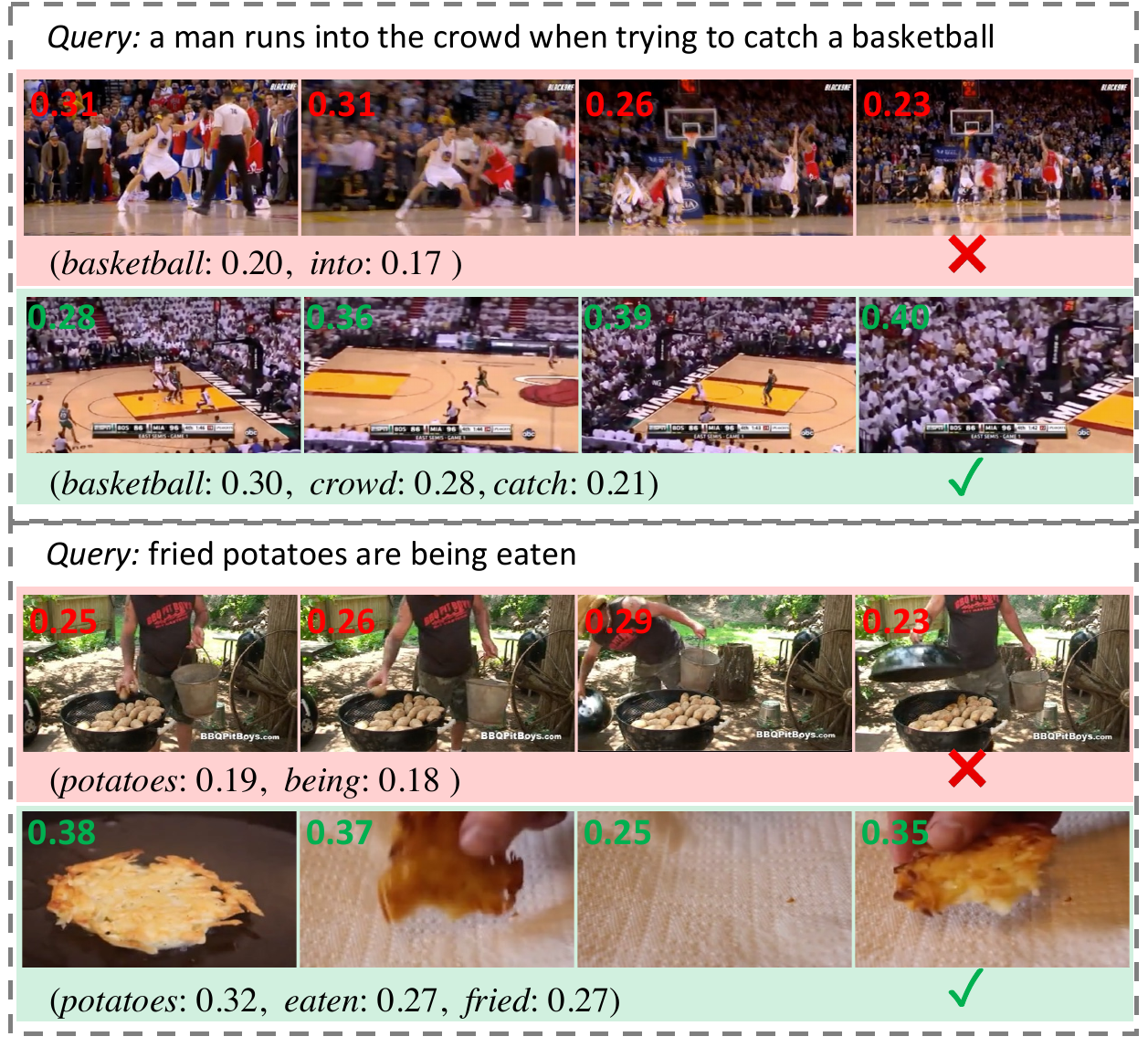}
    \caption{Visualization of TVR results and attention weights for each frame and significant words.
    \textcolor{red}{Red}: incorrect results of the token-wise baseline model. 
    \textcolor{green}{Green}: correct results of our UATVR.}
    \label{fig:uatvr-vis}
\end{figure}

\begin{table}
  \begin{center}
   \scalebox{1.0}{
    \setlength\tabcolsep{3pt}
    \begin{tabular}{l|ccccc}
    \Xhline{0.7pt}
        Method & R@1 & R@5 & R@10 & MdR$\textcolor{green}{\downarrow}$ & MnR$\textcolor{green}{\downarrow}$ \\
        \hline
        CE~\cite{liu2019CE} & 19.8 & 49.0 & 63.8 & 6.0 & - \\
        SUPPORT~\cite{patrick2020supportset} & 28.4 & 60.0 & 72.9 & 4.0 & - \\
        CLIP~\cite{radford2021clip} & 37.0 & 64.1 & 73.8 & 3.0 & - \\
        Frozen~\cite{bain2021frozen} & 33.7 & 64.7 & 76.3 & 3.0 & - \\
        TMVM~\cite{lin2022textadaptive} & 36.7 & 67.4 & 81.3 & 2.5 & - \\
        CLIP4Clip~\cite{luo2021clip4clip} & 45.2 & 75.5 & 84.3 & 2.0 & 10.3 \\
        X-Pool~\cite{gorti2022xpool} & 47.2 & 77.4 & 86.0 & 2.0 & 9.3 \\
        \hline
        \rowcolor{mygray} UATVR(ViT-B32) & 46.0 & 76.3 & 85.1 & 2.0 & 10.4 \\
        \rowcolor{mygray} UATVR(ViT-B16) & \textbf{49.7} & \textbf{79.0} & \textbf{87.3 }& 2.0 & \textbf{8.9} \\
    \Xhline{0.7pt}
      \end{tabular}
  }
  \end{center}
  \caption{\textit{t}2\textit{v} comparisons on the \textbf{MSVD}~\cite{wu2017msvd} dataset.}
  \label{tab:sota-msvd}
  \vspace{-5mm}
\end{table}

\subsection{Qualitative Results}
\label{visualization}
To better understand the impact of additional learnable tokens, we show specific attention weights computed by Eq.\ref{eq:DSA} for each frame and the extra tokens under different text descriptions in \textit{v}2\textit{t} retrieval.
As shown in Fig.~\ref{fig:extra-token-vis}, our model shows higher weights on text-related frames, such as ``\textit{some people are dancing}" and ``\textit{a man discusses his choreography of a play}", resulting in quite a contrary frame attention in the second video.
Moreover, our additional tokens assign higher attention scores to more accurate texts, benefiting text-depended video semantic aggregation during cross-modal matching.
In Fig.~\ref{fig:uatvr-vis}, we show correct UATVR \textit{t}2\textit{v} retrieval results compared to the token-wise baseline. Attention weights for each frame and the most significant words are highlighted.
Due to the matching uncertainty, token-wise baseline easily falls into local context matching with specific queries like `\textit{basketball}' and `\textit{potatoes}'. Nevertheless, UATVR retrieves correct videos with multi-grained high-level reasoning, showing advance in recognizing subtle clues and global semantics simultaneously.
We show further visualization and analysis in the appendix.


\section{Conclusion}
\label{sec:conclution}
In this work, we analyze the uncertain matching problem in existing multi-grained text-video retrieval and propose a novel uncertainty-adaptive matching framework (UATVR) in complementary deterministic and probabilistic views. We model each text-video lookup as a distribution matching procedure by introducing semantic aggregation learnable tokens and distribution-based probabilistic embeddings.
UATVR adaptively addresses the uncertain matching problem and formulates realistic one-to-many text-video correspondences.
Thorough ablation studies and remarkable performance demonstrate the effectiveness of UATVR.
We leave more sophisticated and refined distribution modelling, like a Mixture of Gaussians, as part of future work.

{\small
\bibliographystyle{ieee_fullname}
\bibliography{egbib}
}

\end{document}